\documentclass[letterpaper]{article} 
\usepackage{aaai23}  
\usepackage{times}  
\usepackage{helvet}  
\usepackage{courier}  
\usepackage[hyphens]{url}  
\usepackage{graphicx} 
\usepackage{natbib}  
\usepackage{caption} 
\usepackage{booktabs}
\usepackage{multirow}
\usepackage{xspace}
\usepackage{amsmath,amssymb}
\usepackage[ruled,vlined]{algorithm2e}	
\usepackage{amsthm}
\usepackage{newfloat}
\title{Foresee What You Will Learn:\\Data Augmentation for Domain Generalization in Non-stationary Environment}
\author{
    Qiuhao Zeng\textsuperscript{\rm 1}, Wei Wang\textsuperscript{\rm 1}, Fan Zhou\textsuperscript{\rm 2}, Charles Ling\textsuperscript{\rm 1}, Boyu Wang\textsuperscript{\rm 1}\thanks{Corresponding author: Boyu Wang.}
}
\affiliations{
    \textsuperscript{\rm 1}University of Western Ontario, Canada\\

    \textsuperscript{\rm 2} Beihang University, China\\
    
}

\def\ie{\emph{i.e., }}

\begin{document}

\maketitle

\begin{abstract}
Existing domain generalization aims to learn a generalizable model to perform well even on unseen domains. For many real-world machine learning applications, the data distribution often shifts gradually along domain indices. For example, a self-driving car with a vision system drives from dawn to dusk, with the sky darkening gradually. Therefore, the system must be able to adapt to changes in ambient illumination and continue to drive safely on the road. In this paper, we formulate such problems as Evolving Domain Generalization, where a model aims to generalize well on a target domain by discovering and leveraging the evolving pattern of the environment. We then propose Directional Domain Augmentation (DDA), which simulates the unseen target features by mapping source data as augmentations through a \emph{domain transformer}. Specifically, we formulate DDA as a bi-level optimization problem and solve it through a novel meta-learning approach in the representation space. We evaluate the proposed method on both synthetic datasets and real-world datasets, and empirical results show that our approach can outperform other existing methods. 
\end{abstract}

\section{Introduction}
One common assumption in conventional machine learning methods is that the training and test data are sampled from the same distribution. However, in many real-world problems, this assumption does not hold, and the data distribution can shift in changing environments. 
Consequently, a model learned from training data often fails to generalize well on the data sampled from a shifting distribution, especially when the target data is not accessible. To address the problem of domain shift, \emph{domain generalization} (DG) is proposed to train a model with source domains that can generalize to unseen target domains.

Most existing DG methods aim to extract domain-invariant features by either statistical distance minimization~\cite{muandet2013domain, albuquerque2019generalizing, shui2022benefits,zhou2021domain} or adversarial learning~\cite{li2018domain,volpi2018generalizing,wjd_37}, implicitly assuming that all the domains are independently sampled from a static environment~\cite{muandet2013domain,Arjovsky2019InvariantRM,Sagawa2019DistributionallyRN}. These methods may collapse when the learning tasks are collected from a non-stationary environment. For example, environmental changes due to illumination, seasons, or weather conditions can pose significant challenges for an outdoor robot equipped with vision systems~\cite{Wulfmeier2018IncrementalAD,hoffman2014continuous,lampert2015predicting}. Since the evolving patterns are not taken into account, the existing popular DG methods are not able to handle such problems properly~\cite{muandet2013domain,li2018domain,wjd_88}.



\begin{figure}[t]
    \centering
    \includegraphics[width=0.7\columnwidth]{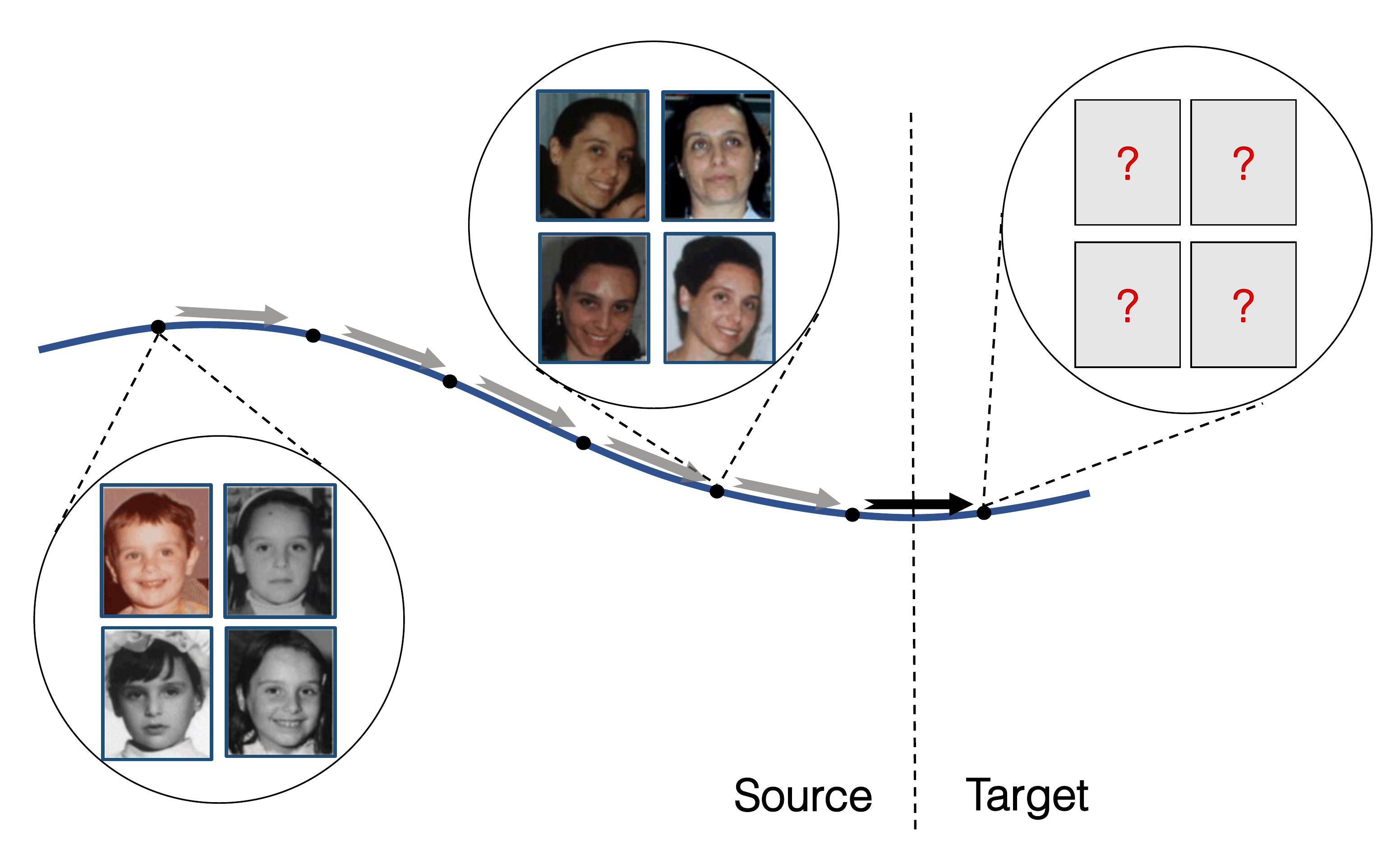}
    \caption{The data distribution shifts along a smoothing curve in a low-dimensional manifold: The appearance of a female changes over years. We capture the data evolving direction (grey arrow) between observed source domains and predict the evolving direction (black arrow) towards the unseen domain beyond the last observed distribution.}
    \label{fig:manifold}
\end{figure}



To alleviate the aforementioned issues, one can take advantage of data shift patterns in non-stationary environments. For example, when deploying a face recognition system to search for a missing person, the system may be built only on photos from childhood to adolescence that were taken decades ago. In such a scenario, the system can benefit from modelling the domain shift along the age to predict a person's current appearance (Fig.~\ref{fig:manifold}). In this work, we address this problem under the \emph{evolving domain generalization} (EDG) scenario~\cite{nasery2021training,qin2022generalizing}, where the source domains are sampled from a changing environment, and the objective is to train a model that generalizes well on an unseen target domain by capturing and leveraging the evolving pattern of the environment.

\begin{figure*}[hbt!] 
    \centering
    \includegraphics[width=1.8\columnwidth]{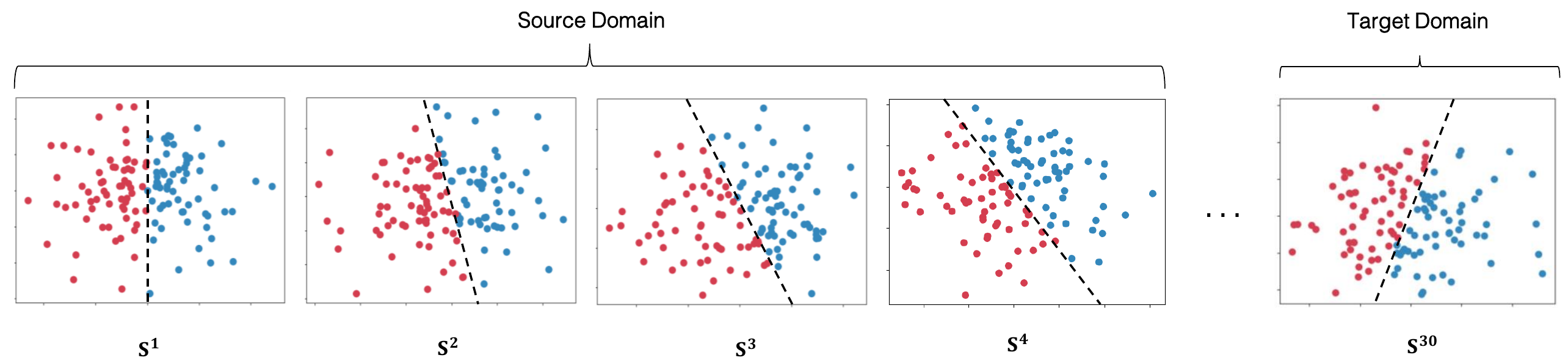}
    \caption{The Rotated Gaussian datasets (dashed lines are the ground truth of decision boundaries)}
    \label{fig:Rotated gaussian}
\end{figure*}




To this end, we propose \emph{directional domain augmentation} (\textbf{DDA}) for EDG. First, we generate augmented features along the direction of the domain shift, such that the augmentations can mimic the next unobserved target domain feature. To achieve this, we design an attention-based \emph{domain transformer} to capture the evolving pattern between consecutive domains by leveraging the power of the attention mechanism to capture the temporal pattern~\cite{girdhar2021anticipative,Vaswani2017AttentionIA} for predicting the feature of future unseen domains. Furthermore, we demonstrate in an illustrative example (Sec.~\ref{sec:motivation example}) that the training process can be formulated as a bi-level programming problem that allows us to effectively capture and leverage the domain shift patterns via a meta-learning scheme.

\noindent To summarize, the contribution of our work is trifold:
\begin{itemize}
    \item[\textbf{1.}] Our work provides a framework to mitigate the impact of lacking data from the target domain under non-stationary environments with an attention-based domain transformer. We show that the optimal domain transformer can generate augmentations whose decision boundaries are aligned with target data without any explicit distribution divergence loss.
    \item[\textbf{2.}] We formulate the training process as a bi-level optimization problem with meta-learning. We also demonstrate that the meta-parameter of the shared classifier could be effectively adapted to the unseen target domain. Our analysis then leads to a novel algorithm, namely~\emph{directional domain augmentation} (\textbf{DDA}), for the EDG problems, which can capture the evolving patterns of domains and predict the future feature effectively.
    \item[\textbf{3.}] We evaluate the algorithm with both synthetic and real-world datasets showing that DDA improves the performance over the state-of-the-art DG algorithms for the EDG problem.
\end{itemize}

\section{Preliminary}\label{sec:preliminary}

\subsection{Problem Setup} 
\label{sec:setup}

Let $\mathcal{D}_t$ be the probability distribution that characterizes $t$-th domain in Evolving Domain Generalization (EDG), and $S_t = \{(x_i^t,y_i^t)\}_{i=1}^{n_t}$ is a set of $n_t$ instances drawn from $\mathcal{D}_t$, where $x_i^t \in\mathcal{X}$ is the $i$-th data point in the $t$-th domain, and $y_i^t\in\mathcal{Y}$ is its label. For every instance, we encode it with a feature extractor $\phi:\mathcal{X} \rightarrow \mathcal{Z}$, and we obtain the embedded instance $z^t_i\in\mathcal{Z}$ by $z^t_i=\phi(x^t_i)$. The goal of EDG is to learn a robust and generalizable model from $T$ source domains by capturing and leveraging the evolving pattern so that it can perform well on the unseen target domain $\mathcal{D}_{T+1}$. 

To this end, we propose a generative approach to EDG which simulates the features for the target domain $\mathcal{D}_{T+1}$ by learning a {domain transformer} $\psi: \mathcal{Z} \rightarrow \mathcal{Z}$. Intuitively, given the data set of the $t$-th domain $S_t$, if $\psi$ can properly capture the evolving pattern, a predictive model $\tilde{h}^*_{t+1}$ trained on the simulated data set $\tilde{S}_{t+1} = \{(\tilde{z}_i^{t+1}, y_{i}^t)\}_{i=1}^{n_t}$, where $\tilde{z}_i^{t+1} = \psi(z_i^t)$, should perform well on $S_{t+1}$. Likewise, a model ${h}^*_{t+1}$ trained on the real data set $S_{t+1}$ should also performs well on $\tilde{S}_{t+1}$. Note that we implicitly assume that the evolving pattern is consistent across all consecutive domains (\emph{i.e.}, $\forall t$, $\psi$ can map instances from $t$ to $t+1$-th domain), which is reasonable in real-world applications. Otherwise, it is impossible to capture the evolving pattern if the environment varies arbitrarily (\emph{e.g.}, it is extremely challenging to predict the stock market tomorrow).  

\begin{figure*}[h]
    \centering
    \includegraphics[width=2.1\columnwidth]{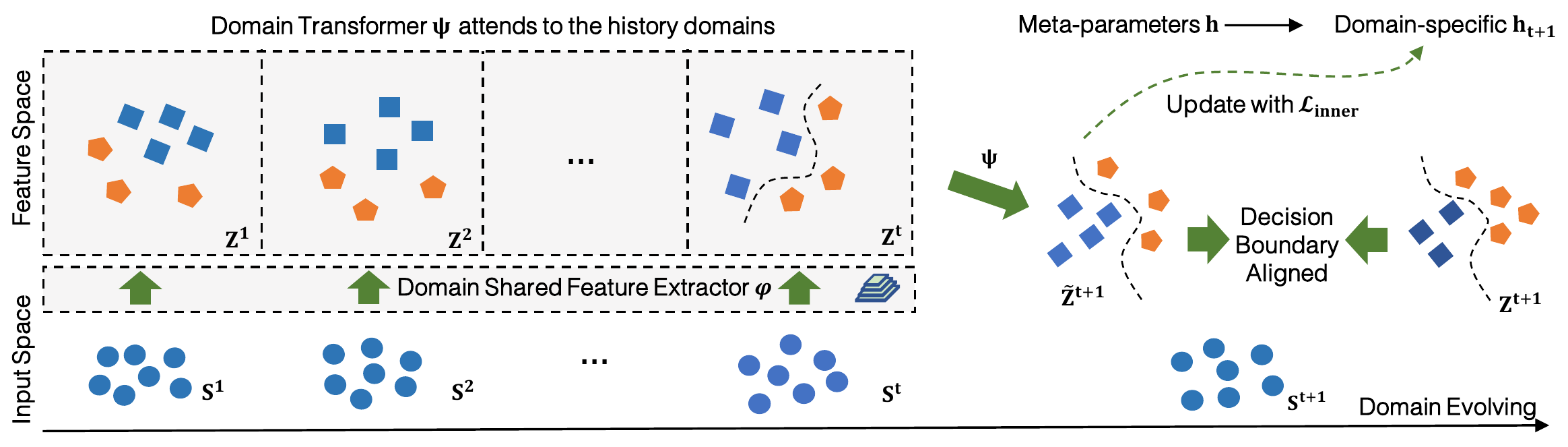}
    \caption{Illustration of the proposed Directional Domain-Augmentation (DDA) model. The domain transformer generates the augmented features in the direction of the unseen domain. With the bi-level optimization, the decision boundary of the augmented embeddings and the embeddings from the unseen domain gets aligned. The meta-parameters are then updated with the softened version classification loss on the augmented features.}
    \label{fig:model}
\end{figure*}

\subsection{An Illustrative Example}
\label{sec:motivation example}
As an illustrative example, we consider the rotated Gaussian data sets as shown in Figure~\ref{fig:Rotated gaussian}, where the instances for binary classification are generated by a $d$-dimensional Gaussian distribution with a mean of $0$, and the decision boundary for each next domain is rotated by 12 degrees counterclockwise. In this example, the domain transformer $\psi$ can be characterized by a rotation matrix: $\psi\in \mathbb{R}^{d\times d}$, and the transform process from $t$ to $t+1$ is simply given by $\tilde{x}_i^{t+1} = \psi x_i^t$ (here we apply an identity mapping as a featurizer $\phi$, so $z_i^t=x_i^t$).  

As analyzed in Section~\ref{sec:setup}, the decision boundaries trained on $S_{t+1}$ and $\tilde{S}_{t+1}$ should be well aligned if $\psi$ can capture the rotation pattern. Thus, we aim to learn $\psi$ in a way such that, for any two consecutive domains, a linear classifier $h_{t+1}^* \in \mathbb{R}^d$ trained on $S_{t+1}$ also performs well on $\tilde{S}_{t+1}$, leading to the following bi-level optimization problem: 


\begin{align}
      &\min_{\psi} \  \|Y^t - \tilde{X}^{t+1} h_{t+1}^*\|_2^2\label{eqn:find psi} \\ 
      &\text{s.t.} \quad   h_{t+1}^* =\; \underset{h}{\text{argmin}}\,\|Y^{t+1} - X^{t+1} h\|_2^2   \label{eqn:subject to} \\
      & \qquad \forall t \in \{1,\cdots, T-1\} \nonumber
\end{align} 
where $Y^t=[y^t_1,\dots, y^t_{n_t}]^\mathsf{T}$, $\tilde{X}^{t+1}=[\tilde{x}^{t+1}_1,\dots, \tilde{x}^{t+1}_{n_t}]^\mathsf{T}$, ${X}^{t+1}=[{x}^{t+1}_1,\dots, {x}^{t+1}_{n_t}]^\mathsf{T}$, and $\mathsf{T}$ is the transpose operator. Here, we adopt the squared loss for classification in order to obtain an analytical solution, which also corresponds to Fisher's linear discriminant~\cite{bishop2006pattern}.

Setting the derivative with respect to $h_{t+1}$ to zero for Eqn.~\ref{eqn:subject to}, we obtain the optimal solution $h_{t+1}^{*}=[(X^{t+1})^\mathsf{T}X^{t+1}]^{-1}(X^{t+1})^\mathsf{T} Y^{t+1}$. Similarly, setting the derivative with respect to $\psi$ to zero for Eqn.~\ref{eqn:find psi}, we obtain the following equation for $\psi$ (detailed derivation can be found in Section~A of the supplementary materials):
\begin{align}
    A^t\text{vec}(\psi) = B^t,
    \label{eqn:rearange}
\end{align}
where function vec$(\cdot)$ is the vectorization operator, $A^t = \bigl( h_{t+1}^*  (h_{t+1}^*)^\mathsf{T}\bigl) \otimes \bigl({X^t}^\mathsf{T} X^t\bigl)$, $B^t = \text{vec}({X^t}^\mathsf{T} Y^t {h_{t+1}^*}^\mathsf{T})$, and $\otimes$ is the Kronecker product operator.

As the rotation is consistent across all domains, Eqn.~\ref{eqn:rearange} holds for any $t=1,\dots, T-1$, which gives
\begin{equation}\label{eqn:rotate}
    [(A^1)^\mathsf{T},\ldots, (A^{T-1})^{\mathsf{T}}]^\mathsf{T} \text{vec}(\psi)= [(B^1)^\mathsf{T}, \ldots, (B^{T-1})^{\mathsf{T}}]^\mathsf{T}.
\end{equation}
As $\text{rank}(A^t)=d$, Eqn.~\ref{eqn:rotate} will be well-posed when $d\le T-1$. Then, for the rotated Gaussian shown in Fig.~\ref{fig:Rotated gaussian}, the domain transformer $\psi^*$ learned by solving Eqn.~\ref{eqn:rotate} is given by
\begin{equation*}
    \psi^*=\left[\begin{array}{cc}
       0.9824  & 0.2031 \\
      -0.2107   & 0.9720\\
    \end{array}\right]  \approx
    \left[\begin{array}{cc}
       \cos{12^\circ}  & \sin{12^\circ} \\
      -\sin{12^\circ}   & \cos{12^\circ}\\
    \end{array}\right],
\end{equation*}
which is very close to the ground-truth rotation matrix.

We denote this method as the linear DDA (LDDA), and its classification accuracy on the unseen target domain is shown in Table~\ref{Table:dda-heuristic} (see Section~\ref{sec:settings} for the details of other baseline algorithms), from which it can be observed that  LDDA achieves the best performance.

\begin{table}[]
\setlength{\tabcolsep}{3.5pt}
\begin{center}
\begin{small}
\begin{sc}
\resizebox{1\columnwidth}{!}{
\LARGE
\begin{tabular}{ccccc}
\toprule
ERM & CIDA & EAML & LSSAE & LDDA\\
\midrule

59.2 $\pm$ 1.1 & 50.5 $\pm$ 1.5  & 61.0 $\pm$ 2.8 & 88.4 $\pm$ 0.8& 94.6 $\pm$ 0\\
\bottomrule
\end{tabular}
}

\end{sc}
\end{small}
\end{center}
\caption{Experiment Results (accuracy \%) on Synthetic Rotated Gaussian dataset}
\label{Table:dda-heuristic}
\end{table}


\section{Method}
\label{sec:method}

In the rotated Gaussian example, LDDA aims to learn a domain transformer $\psi$ to capture the evolving patterns of the environment, which motivates an effective solution to the EDG problem by designing a bi-level optimization problem. Note that $\psi$ of LDDA in the illustrative example is assumed as a linear mapping in order to obtain an analytical solution and is only applicable to low-dimensional cases (\emph{i.e.}, solving Eqn.\ref{eqn:rotate} requires $d\le T-1$). We will illustrate our DDA framework in this section which extends to the general cases, including the non-linear cases.

\subsection{Method Overview }
In this section, we extend the proposed method to deep models by learning the domain transformer in the representation space and solving the bi-level optimization problem through a novel meta-learning scheme. Specifically, the proposed model consists of three components: a feature extractor $\phi$, a domain transformer $\psi$, and a classification model $h$, which are parameterized by $\theta_\phi$, $\theta_\psi$, and $\theta_h$, respectively. 

DDA's overall design is illustrated in Fig.~\ref{fig:model}. The sampled inputs first get projected into feature space by $\phi$. $\psi$ generates the augmentations to mimic the data from the next unseen domain by leveraging the evolving pattern. $h$ takes fast adaptations on augmentations and the optimized $h$ will perform well in the classification task of the next domain.

\subsection{Domain Transformer}
\label{subsec:domain transformer}



To capture the evolving patterns, the domain transformer $\psi$ is designed to generate augmentations by transforming the features from historical domains into the next domain. In Section~\ref{sec:preliminary}, only one preceding domain is utilized to simulate the next domain in a point-wise manner (\ie{$\psi$ only maps $z_i^t$ to $\tilde{z}_i^{t+1}$}). In order to take full advantage of the source data, we design $\psi$ with an attention module, taking its strength to extract sequential information~\cite{Vaswani2017AttentionIA,zeng2020learning}, which allows the information to propagate over the evolving domains. Consequently, it can leverage the data from all historical source domains to effectively capture evolving patterns and simulate the target data. 


Specifically, we first calculate the similarity score $s_{i,j}^{t,t'}$ between $i$-th sample from domain $t$ and $j$-th sample from a historical domain $t'$,    
\begin{align}
\label{Eqn:scores}
    s_{i,j}^{t,t'}&=  \frac{\psi_q(z_i^t)\psi_k(z_j^{t'})}{\sqrt{d}}
\end{align}
where $1\le i,j \le B$, $1\le t' \le t \le T$, $B$ is the batch size, $\psi_q(\cdot)$ and $\psi_k(\cdot)$ denote the  transformation that maps input features into the query and key embeddings, and $z_j^{t'}$ is a sampled feature from the historical domains in a batch. The similarity score $s_{i,j}^{t,t'}$ is normalized by the dimension of each transformed embedding to avoid small gradients caused by subsequent \emph{softmax} functions~\cite{Vaswani2017AttentionIA}. It measures how close the current sample is to the historical domain samples. Therefore, $\psi$ will attend to the most relevant samples by reviewing the entire domain evolving history. Then, the directional transform augmentations $\tilde{z}_i^{t+1}$ can be obtained by a weighted sum of embedded values in the history domain plus the output of a skip-connection network:
\begin{equation}
    \tilde{z}_i^{t+1} =\sum\limits_{t'=1}^{t}\sum\limits_{j=1}^{B} \frac{\exp{s_{i,j}^{t,t'}}}{\sum\limits_{t''=1}^{t}\sum\limits_{k=1}^{B}\exp{s_{i,k}^{t,t''}}} \psi_v(z_j^{t'}) + \psi_{\text{sc}}(z_i^t)
    \label{eqn:domain transformer}
\end{equation}
where $1\le i,j,k \le B$, $1\le t',t'' \le t \le T$, $\psi_v(\cdot)$ denotes the transform that maps input embeddings into value embeddings. $\psi_{\text{sc}}$ is a skip-connection network to help stabilize the learning~\cite{zhang2018image}. In our case, it also helps to preserve the instance-level information. Thus, the domain transformer $\psi = \{\psi_k,\psi_q,\psi_v,\psi_{\text{sc}}\}$. We name augmentation $\tilde{z}_i^{t+1}$ \emph{directional transform augmentation}, as it is generated according to the domain-evolving direction and transformed from the samples of the historical domains.  It is noted that a sample input $z^t_i$ is taken as a query, and the samples from history domains are taken as keys and values. The domain transformer aggregates information across domains~\cite{xu2021cdtrans} and generates $\tilde{z}^{t+1}_i$ in a way such that its decision boundary can also correctly classify the features ${z}^{t+1}_i$ from the next domain.


\scalebox{0.1}{}{
\begin{algorithm}[t]

    \SetKw{Input}{Input:}\SetKwInOut{Output}{output}
    \SetKw{Define}{define}\SetKw{As}{as}
    \caption{Directional Domain Augmentation}
    \Input{The feature extractor $\phi$, the domain-shared classifier $h$, the domain transformer $\psi$, the learning rate of the inner loop $\alpha$, the learning rate of the outer loop $\beta$ and the batch size $B$.}\\
    Initialize $\theta$ ($\theta=\left\{\theta_{h}, \theta_{\psi}, \theta_\phi\right\}$ ) \\
    \For{sampled mini-batch $\left\{\{x_i^t, y_i^t\}_{i=1}^B\right\}_{t=1}^T$}{
        Calculate instance features $\{Z^t\}_{t=1}^T$\;
        $\mathcal{L}=0$ \tcp*{Initialize the loss for this episode}
        \For{$t=1$ to $T-1$}{
            for every sample $i$ in each domain $t$:\\
            $\qquad\tilde{z}^{t+1}_i = \psi(z^t_i|\{Z^{t'}\}_{t'=1}^{t-1})$ in Eqn.~\ref{eqn:domain transformer}\;
            \For{k=1 to num. of inner-loop steps}{
            Calculate the inner loss $\mathcal{L}_{\text{inner}}$ in Eqn.~\ref{Eqn:Inner}\\
                $\theta_{h_{t+1}} = \theta_h - \alpha \nabla_{\theta_{h}}\mathcal{L}_{\text{inner}} $\\
            }
            Calculate the outer loss $\mathcal{L}_{\text{outer}}$ in Eqn.~\ref{Eqn:out loss}\\
            $\mathcal{L}=\mathcal{L}+\mathcal{L}_{\text{outer}}$\\
                
        }
        Update $\theta \leftarrow \theta -\beta \nabla_\theta \mathcal{L}$ \; 
    }
    \KwRet{trained model parameters $\theta$ } 
        \label{Algo}
    
\end{algorithm}}

\subsection{Bi-level Optimization with Meta-learning}
\label{sec:bilevel}
As shown in Eqn.~\ref{eqn:find psi} and Eqn.~\ref{eqn:subject to}, we aim to optimize $\psi$ through a bi-level optimization scheme so that $\tilde{S}^{t+1}$ and $S^{t+1}$ can share the same predictive model $h_{t+1}^*$. One issue with this scheme is that each $h_{t+1}^*$ in the inner loop (\ie{Eqn.~\ref{eqn:subject to}}) is only learned from a single domain, which may lead to the overfitting problem. In order to take advantage of the transferred knowledge from all the other source domains, instead of learning $h_{t+1}^*$ for each domain individually, we learn a good initialization $\theta_h$ that is shared across all the domains, and each domain-specific classifier, parameterized by $\theta_{h_t}$, can be learned from $S_t$ via \emph{fast adaptations}~\cite{finn2017model}.


Therefore, learning $\theta_\psi$ and $\theta_h$ can be seamlessly integrated into a single bi-level optimization problem that can be solved by meta-learning, resulting in more effective use of data.  Specifically, we apply the episodic training scheme in~\cite{finn2017model}, which consists of two steps: inner-loop updates and outer-loop updates. The training protocol is shown in Algorithm~\ref{Algo}. 


In each episode, we sample $B$ data points of each domain from domain $1$ to domain $T$, yielding $\{\{x_i^t, y_i^t\}_{i=1}^B\}_{t=1}^T$. Let $Z^t = \{z_i^t\}_{i=1}^B$ and $\tilde{Z}^{t+1}=\{\tilde{z}_i^{t+1}\}_{i=1}^B$, respectively, be features of sampled batch instances from the $t$-th domain and its directional transform augmentations. We randomly select two consecutive domains, domain $t$ and domain $t+1$. Then, $\theta_h$ is learned with the loss $\mathcal{L}_{\text{inner}}$ on $\tilde{Z}^{t+1}$ in the inner loop:
\begin{align}
\label{Eqn:Inner}
        &\mathcal{L}_{\text{inner}} (S_t;\theta_h, \theta_{\psi},\theta_{\phi})  = \frac{1}{B}\sum_{i=1}^{B} [ \lambda \cdot \mathcal{L}_{cls}(y_{t}^i, h(\tilde{z}_{t+1}^i)) \nonumber \\
        &+ (1-\lambda) \mathcal{D}_{KL}(\sigma(h_{t}(z_{t}^i))/\tau_{\text{temp}}|| \sigma(h(\tilde{z}_{t+1}^i))/\tau_{\text{temp}})]
\end{align}
where $\mathcal{L}_{cls}$ is the cross-entropy loss,  $\sigma$ is the softmax function, and $\lambda$ is a trade-off parameter.  $\mathcal{D}_{KL}(\cdot||\cdot)$ is the Kullback-Leibler (KL) divergence, which is adopted as a distillation loss~\cite{Hinton2015DistillingTK}. It can be  regarded as a softened softmax at a temperature $\tau_{\text{temp}}$ and able to reserve the instance semantics. Then, the domain-specific classifier $\theta_{h_{t+1}}$ is given by
\begin{equation}
\label{Eqn:inner-update}
    \theta_{h_{t+1}} = \theta_h - \alpha \nabla_{\theta_{h}}\mathcal{L}_{\text{inner}} (S_t;\theta_{h}, \theta_{\psi}, \theta_\phi)
\end{equation}
where $\alpha$ is the inner-loop learning rate. $\theta_{h_{t+1}}$ is the classifier optimized with $\tilde{Z}^{t+1}$, which is shared with the target domain's instances. Therefore, in the outer loop of each episode, $h_{t+1}$ is evaluated on $Z^{t+1}$, and the corresponding loss function  $\mathcal{L}_{\text{outer}}$ is given by
\begin{align}\label{Eqn:out loss}
    &\mathcal{L}_{\text{outer}} (S_t, S_{t+1};\theta_{h_{t+1}}, \theta_{\psi}, \theta_\phi)\\
    &\hspace{16pt}=\frac{1}{B}\sum_{i=1}^{B} \mathcal{L}_{cls}(y_{t+1}^i, h_{t+1}(z_{t+1}^i|\theta_{h_{t+1}}))\nonumber
\end{align}


Then, overall DDA parameters is updated by
\begin{equation}
    \theta \leftarrow \theta -\beta \nabla_\theta \mathcal{L}_{\text{outer}}, \qquad \theta=\left\{\theta_{h}, \theta_{\psi}, \theta_\phi\right\}
\end{equation}where $\beta$ is the outer-loop learning rate.

In the inference stage, we first simulate a set of the feature augmentations of size $N$: $\tilde{Z}^{T+1}=\{\tilde{z}^{T+1}_i\}_{i=1}^N$ from historical source domain features, and obtain the parameter $\theta_{h_{T+1}}$ for the target classifier on $\tilde{Z}^{T+1}$ via fast adaptation from $\theta_h$.   

\begin{table*}[ht!]
\setlength{\tabcolsep}{3.5pt}

\begin{center}
\begin{small}
\begin{sc}
\resizebox{2\columnwidth}{!}{
\LARGE
\begin{tabular}{ccccccccc}
\toprule
\textbf{Dataset} & Sine &Rotated Gaussian & Portrait & Rotating MNIST &	Forest Cover & Ocular Disease & CalTran & Average\\
\midrule
ERM & 56.3 $\pm$ 1.2 & 59.2 $\pm$ 1.1 & 90.3 $\pm$ 0.1 &78.2 $\pm$ 0.2 & 59.8 $\pm$ 0.2&  71.2 $\pm$ 0.3 & 96.6 $\pm$ 0.7 & 73.1 \\
GroupDRO& 62.6 $\pm$ 1.5& 80.8 $\pm$ 3.4 & 92.6 $\pm$ 0.2 & 79.1 $\pm$ 0.1 & 58.9 $\pm$ 0.5&  71.3 $\pm$ 0.2  & 96.6  $\pm$ 0.4 & 77.4\\
IRM & 51.1 $\pm$ 2.3 & 72.0 $\pm$ 2.2 & 91.3 $\pm$ 0.4 & 79.2 $\pm$ 0.3 & 58.8 $\pm$ 0.8& 69.8 $\pm$ 0.4 & 94.9 $\pm$ 1.2 & 73.9\\
MMD & 54.7 $\pm$ 4.7 & 56.8 $\pm$ 1.3 & 92.0 $\pm$ 0.2 &77.4 $\pm$ 0.0  & 59.0 $\pm$ 0.3& 67.7 $\pm$ 0.3 & 97.4 $\pm$ 0.2  & 72.1\\ 
CORAL & 54.7 $\pm$ 5.4 &56.8 $\pm$ 1.1& 91.3 $\pm$ 0.2 & 78.9 $\pm$ 0.1& 62.0 $\pm$ 1.1& 67.8 $\pm$ 0.5& 96.6 $\pm$ 0.5 & 72.6\\
MTL  &54.2 $\pm$ 3.2& 56.4 $\pm$ 1.4&92.0 $\pm$ 0.1 & 79.0 $\pm$ 0.2&60.4 $\pm$ 0.7&71.3 $\pm$ 0.4& 97.5 $\pm$ 0.3 & 73.0\\
MLDG  & 54.7 $\pm$ 2.5& 53.6 $\pm$ 2.1 & 91.5 $\pm$ 1.1 &82.8 $\pm$ 0.2& 60.9 $\pm$ 0.7& 72.3 $\pm$ 0.3 & 97.3 $\pm$ 0.2 & 73.3 \\
SagNet & 51.1 $\pm$ 3.1& 52.0 $\pm$ 1.8 & 92.7 $\pm$ 0.2 & 80.9 $\pm$ 0.1& 62.1 $\pm$ 2.0& 69.3 $\pm$ 0.4 & 97.2 $\pm$ 0.1 & 72.2\\
SelfReg  & 55.8 $\pm$ 1.7& 54.4 $\pm$ 1.0 & 90.6 $\pm$ 0.3 & 81.8 $\pm$ 0.5 & 60.1 $\pm$ 0.6 & 65.2 $\pm$ 0.1 & 96.5 $\pm$ 0.4 & 72.1 \\
DAML & 52.6 $\pm$ 0.7 & 62.3 $\pm$ 1.3 & 92.7 $\pm$ 0.3 & 84.0 $\pm$ 0.4 & 61.3 $\pm$ 0.6 & 71.2 $\pm$ 0.2 & 95.9 $\pm$ 0.3 & 74.3\\
CIDA & 65.1 $\pm$ 3.7  & 50.5 $\pm$ 1.5 & 92.3 $\pm$ 0.4 & 83.6 $\pm$ 1.2 & 60.5 $\pm$ 0.9 & 71.4 $\pm$ 0.3 & 97.1 $\pm$ 0.7 & 74.4 \\
EAML & 49.0 $\pm$ 0.7& 61.0 $\pm$ 2.8& 90.1 $\pm$ 0.4 &82.6 $\pm$ 0.2 &  60.8 $\pm$ 1.4 & 71.7 $\pm$ 0.6 & 96.5 $\pm$ 0.6 & 73.1 \\
LSSAE & 63.2 $\pm$ 1.5 & 88.4 $\pm$ 0.8 & 93.1 $\pm$ 0.3& 84.7 $\pm$ 0.3& 63.2 $\pm$ 0.4 & 72.4 $\pm$ 0.4 & 97.2 $\pm$ 1.0 & 80.3\\ 
GI & 66.8 $\pm$ 0.7 & 85.1 $\pm$ 0.5 &93.7 $\pm$ 0.2 & 83.4 $\pm$ 0.7 &   63.6 $\pm$ 0.4 & 73.1 $\pm$ 0.2 & 98.2 $\pm$ 0.8 & 80.6\\
\textbf{Our Method} &\textbf{98.4 $\pm$ 0.9} &\textbf{99.6 $\pm$ 0.6} &\textbf{94.9 $\pm$ 0.1} & \textbf{86.2 $\pm$ 0.3}& \textbf{65.3 $\pm$ 0.5} & \textbf{74.1 $\pm$ 0.1} & \textbf{98.3 $\pm$ 0.4} & \textbf{88.1}\\

\bottomrule
\end{tabular}}

\end{sc}
\end{small}
\end{center}
\caption{Experiment Results (accuracy \%) on Synthetic Dataset and Real-World Datasets among different methods}
\label{Table:real-world}
\end{table*}


\section{Related Work}

\noindent
\paragraph{Domain Generalization (DG)} Distribution matching is one predominant approach in domain generalization, where domain-invariant representation learning~\cite{wjd_78} is intuitive and has been extensively studied. \cite{wjd_104} proposed to ensemble a unified model with generalization capability. Meta-learning has also been investigated for generalization \cite{li2018learning,li2020sequential,balaji2018metareg,wjd_14}. It is common sense that existing DG methods can not handle extrapolation well \cite{domainbed,nguyen2021domain}, which makes it not suitable for our problem setup. In the early stage, the researchers mainly focused on aligning feature marginal distributions\cite{wjd_78}, which has been proved not enough in case there exists concept-shift across domains. As a consequence, many recent works proposed to align the joint distributions\cite{li2021learning,nguyen2021domain}. However, joint distribution alignment is much harder than marginal distribution alignment, which results in the introduction of varieties of techniques such as information theory\cite{li2021learning}.

\noindent
\paragraph{Data augmentation} Conventional data augmentation operations include cropping, flipping, rotation, scaling, and nosing. Data augmentation has been applied to improve the generalization capability of DG models. Besides conventional data augmentation methods, there are also a large number of generating-based methods~\cite{wjd_38,wjd_120} trying to generate all new instances. For example,~\cite{wjd_37} trains a transformation network for data augmentation.

Recent approaches~\cite{Volpi2019AddressingMV,Shi2020TowardsUR,Zhou2020LearningTG,Zhou2021DomainGW} have studied the data augmentation methods in DG by generating either augmented samples or intermediate embeddings to improve the generalization performance on the unseen domains. However, the domain shift patterns are absent in these kinds of approaches making the methods lack the ability to learn the non-stationary evolving patterns. 


\noindent
\paragraph{Evolving Domain Adaptation (EDA) / Evolving Domain Generalization (EDG)} Several existing works have formulated a similar scenario as \emph{evolving domain adaptation}~\cite{hoffman2014continuous,lampert2015predicting,wang2020continuously,Wulfmeier2018IncrementalAD}, where the environment can change in a continuously evolving way. \cite{kumagai2016learning} predicts future classifiers on the basis of variational Bayesian inference by incorporating the vector auto-regressive model to capture the dynamics of a decision boundary. \cite{Wulfmeier2018IncrementalAD,wang2020continuously} learn the representations that are time-invariant using adversarial methods. We emphasize that EDA still has access to unlabeled data from upcoming target domains to help learn evolving patterns, while EDG has no access to the target data at all. 

There are very few works \cite{nasery2021training,qin2022generalizing} tackling EDG problems. \cite{nasery2021training} learns an operator that captures the evolving dynamics of the time-varying data distribution. \cite{qin2022generalizing} proposes a novel probabilistic framework named LSSAE by incorporating variational inference to identify the continuous latent structures of \emph{concept shift} and \emph{covariate shift} under EDG settings. Both two EDG methods design complex Neural-Network structures and did not utilize the decision boundary alignment to mitigate the evolving domain shift, which has been verified as efficient in our illustrated example and the corresponding analysis.

\section{Experiment}
To evaluate our method, we demonstrate our method on several toy datasets including Sine and Rotated Gaussian toy datasets and also on the real-world datasets, including \textbf{Portraits}, \textbf{Cover Type}, \textbf{Ocular Disease} and \textbf{Caltran} (We delegate description of the datasets to the supplementary materials~B). Extensive ablation studies are conducted to show the effectiveness of the meta-learning of our method. 

\subsection{Experiment Settings}\label{sec:settings}

We evaluate the proposed method with the following baselines: (1) \textbf{ERM}~\cite{Vapnik1998StatisticalLT}; (2) \textbf{GroupDRO}~\cite{Sagawa2019DistributionallyRN}; (3) \textbf{IRM}~\cite{Arjovsky2019InvariantRM}; (4) \textbf{CORAL}~\cite{Sun2016DeepCC}; (5) \textbf{MMD}~\cite{li2018domain}; (6) \textbf{MLDG}~\cite{li2018learning}; (7) \textbf{SagNet}~\cite{Nam2021ReducingDG}; (8) \textbf{SelfReg}~\cite{kim2021selfreg}; (9) \textbf{DAML}~\cite{shu2021open}; (10) \textbf{CIDA}~\cite{wang2020continuously}; (11) \textbf{EAML}~\cite{liu2020learning}; (12) \textbf{LSSAE}~\cite{qin2022generalizing}; (13) \textbf{GI}~\cite{nasery2021training}.
All the baselines and experiments were implemented with DomainBed package~\cite{domainbed} under the same settings, which guarantees fair and sufficient comparisons. For all benchmarks, we conduct the leave-one-domain-out evaluation. We train our model on the validation splits of all seen source domains (domain $1$, $2$, ..., $T$) and select the best model on the validation of all source domains. For testing, we evaluate the selected model on all images of the held-out unseen target domain (domain $T+1$).

\begin{figure*}[t] 
    \centering
    \includegraphics[width=2.1\columnwidth]{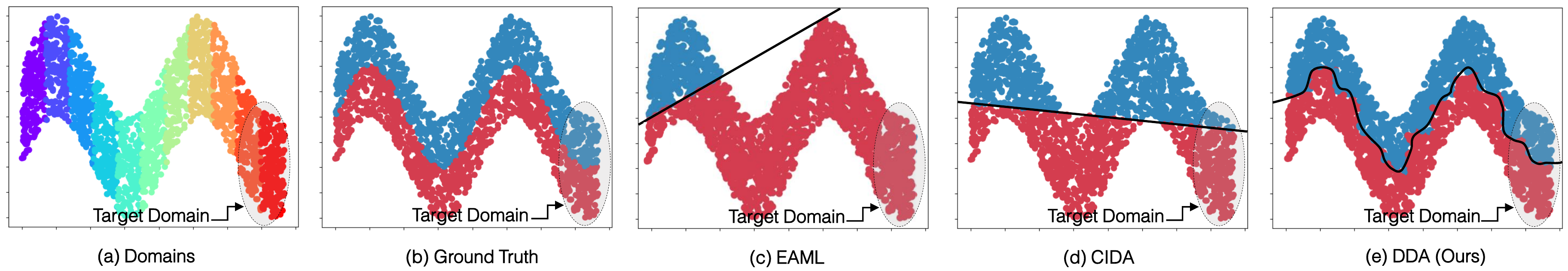}
    
    \caption{Results on the Sine dataset with 11 domains. We set the classification model as a single linear layer which makes this task extremely difficult. The black line is the decision boundary predicted by the model. (a) Domains are indexed by color. The first 10 domains are source domains, marked by purple to orange color. The 11th domain is the test domain, marked by red color and a circle. (b) The ground truth of the decision boundaries separates positive and negative samples. (c) The prediction results of EAML on the source and target domains. (d) The prediction results of CIDA on the source and target domains. (e) The prediction results of DDA on the source and target domains.}
    \label{fig:Sine}
\end{figure*}

\subsection{Evaluation on Synthetic Dataset and Real-World Dataset}

From Table~\ref{Table:real-world} we can see, most algorithms fail on both Sine and rotated Gaussian datasets. Since it is a binary classification task, other methods with about 50$\%$ accuracy are doing random predictions. One reason is that their decision boundaries are static and do not make any adjustments to the direction of the data distribution shift. DDA could successfully capture such shifts and adaptively adjust the decision boundary to fit each domain including target domains. Fig. \ref{fig:Sine} visualize the Sine dataset, which indicates an adaptive model as DDA with domain-specific classifiers can address the EDG problem properly. 

DDA also achieves the best performance on real-world datasets. In the Rotating MNIST dataset, our algorithm could achieve 86.2\% accuracy, which is 1.5\% higher than the second best method among the baselines. Specifically, the results on the Rotating MNIST dataset are the average accuracies under different experimental settings (different number of domain intervals, different total number of domains). On Portrait and Ocular Disease datasets, DDA achieves $94.9\%$ which is $1.2\%$ higher than the best baselines. Ocular Disease contains medical photographs from 5,000 patients that vary with the age of the subjects. Our method improves the performance by $1.0\%$ compared to the second best baseline, achieving $74.1\%$ accuracy. Caltran contains images of traffic taken with stationary cameras over time. Likewise, our method also improves performance by capturing evolving patterns. They show the possibility that our method can be deployed in real-world applications. 

MTL also augments the feature space with the marginal distribution of features. It indicates the superiority of data augmentations on DG problems but MTL fails to capture the domain evolving patterns and improve the performance by generalizing to unknown domains with random directions. Hence, MTL is still worse than DDA. In our experiments, CIDA and EAML can not achieve good performance even with access to target unlabeled data.
The reasons may be that both methods fail to capture the evolving pattern but instead learn domain-invariant encodings. This also shows capturing evolving patterns is critical to solving the EDG problem.

\subsection{Non-stationary Environments with Multiple Target Domains}
In practice, data can be streamed continuously from multiple future domains. Therefore, we also conducted experiments on Rotating MNIST by dividing more domains into target domains in Table~\ref{Table:T+3}. The experimental setup is to have 6 source domains with rotation degrees of $[0^\circ, 15^\circ, 30^\circ,45^\circ,60^\circ,75^\circ]$ and 3 target domains with rotation degrees $[90^\circ,105^\circ,120^\circ]$. The results in Table~\ref{Table:T+3} demonstrates that our algorithm can also achieve better performance than other baselines in future steps. In order to generate augmentations in the $(t+2)$-th domain and the $(t+3)$-th domain, it is necessary to make some modifications to our algorithm. We show the modification details in the supplementary material~E. 


\begin{figure}[t]
    \centering
    \includegraphics[width=8.5cm]{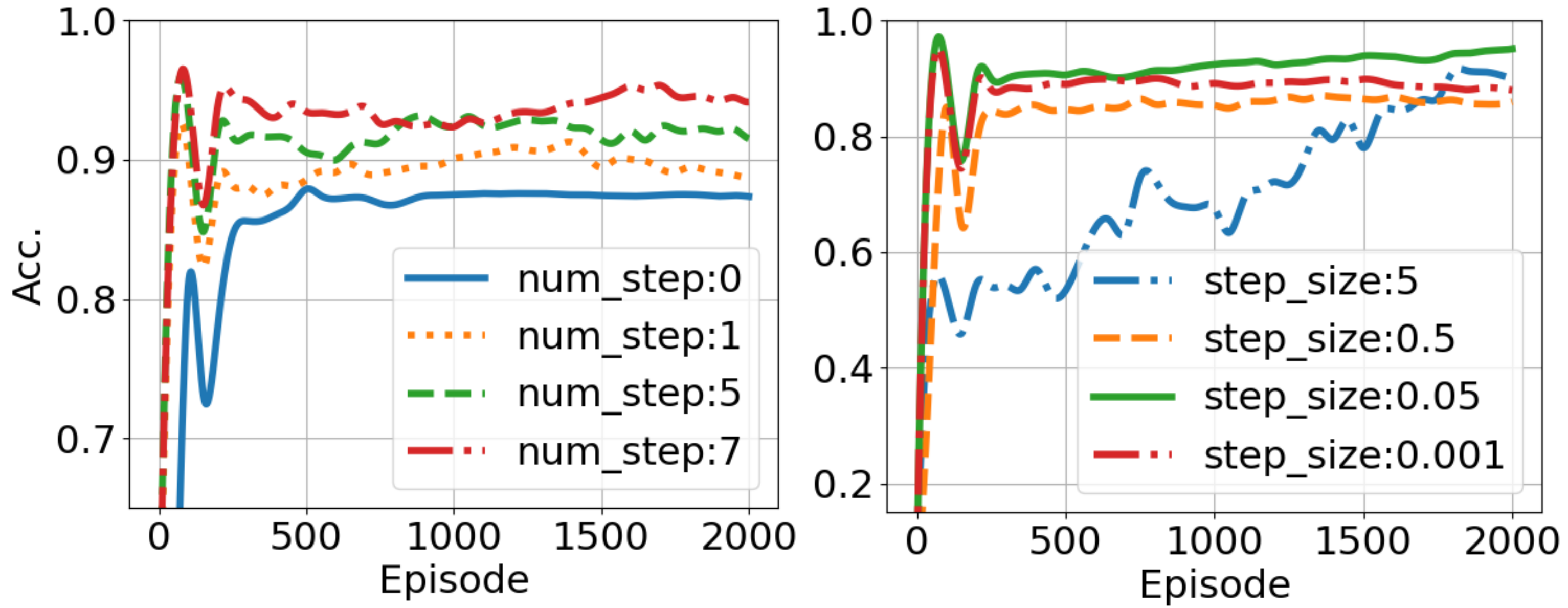}
    \caption{(Left) The convergence trajectory of the test accuracy with the different number of inner meta-updating steps on Portrait Dataset. (Right) The convergence trajectory of the test accuracy with the different step size $\alpha$ of inner meta-updating steps on Portrait Dataset.}
    \label{Fig:sizes & steps}
\end{figure}

\begin{table}[h!]
\setlength{\tabcolsep}{3.5pt}
\begin{center}
\begin{small}
\begin{sc}
\begin{tabular}{cccc}
\toprule
\textbf{domains}& \textbf{T+1}&\textbf{T+2}&\textbf{T+3}\\
\midrule
ERM & 81.0 $\pm$ 0.2 &  56.5 $\pm$ 0.3 &  39.9 $\pm$ 0.3  \\
MLDG & 87.9 $\pm$ 0.3 &  66.1 $\pm$ 0.4  &  45.6 $\pm$ 0.4\\


CIDA & 87.0 $\pm$ 1.3 & 73.5 $\pm$ 1.2  & 48.1 $\pm$ 1.4\\
EAML & 88.6 $\pm$ 0.3 & 72.2 $\pm$ 0.3  & 49.9 $\pm$ 0.4\\
GI & 89.6 $\pm$ 0.2 & 73.6 $\pm$ 0.3 & 52.4 $\pm$ 0.2\\
LSSAE & 88.9 $\pm$ 0.3& 74.2 $\pm$ 0.3 & 51.1 $\pm$ 0.3\\

Our Method & \textbf{92.3 $\pm$ 0.2} & \textbf{77.0 $\pm$ 0.2}  &  \textbf{55.8 $\pm$ 0.3}  \\
\bottomrule
\end{tabular}
\end{sc}
\end{small}
\end{center}
\caption{Experiment Results (accuracy \%) on the Rotating MNIST dataset with multiple target domains}
\label{Table:T+3}
\end{table}

\subsection{Ablation Study}

\textbf{Comparison with different numbers of inner updating steps}
As \cite{finn2017model} points out the number of update steps affects the convergence speed and performance, we test effects of the number of internal update steps in Fig.~\ref{Fig:sizes & steps}: Left. Multiple inner-loop steps will result in more computations. To reduce time complexity, we always update 2 steps on the source domains; meanwhile, we take 1 to 10 inner steps on the directional transform augmentations of the target domain. Results show the performance of DDA improves as the number of inner-loop steps increases. 

The performance drops drastically by setting the number of steps to $0$. With fast adaptations of the inner loop, the accuracy of the accuracy trajectory converges faster if the inner loop step size is set to $0$. The performance drops drastically by setting the number of steps to $0$, in which case the model is not equipped with a classification component parameterized with meta-parameters, but a domain-invariant classifier.

\noindent\textbf{Comparison of step sizes in the inner-loop}
The step size $\alpha$ in Eqn.~\ref{Eqn:inner-update} is a factor related to the distance between domains. As the domain interval between the evolving domains is larger, it requires a larger $\alpha$. 
From Fig.~\ref{Fig:sizes & steps}: Right we can see, setting $\alpha$ as $0.05$ is the best choice for the Portrait dataset.

\noindent\textbf{Different domain interval between domains} In Table~\ref{Table:Interval}, the intervals of rotation degrees between domains are set to 10$^{\circ}$, 20$^{\circ}$, 30$^{\circ}$ and the total number of domains is fixed to 9. Our proposed method outperforms all the baselines. As the domain interval increases, we can see that all methods' performance degrades. This is because, with a larger domain discrepancy caused by the bigger domain interval, the model gets harder to capture the robust representations for classification tasks. Specifically, when the domain interval is 30$^{\circ}$, our method outperforms the best baseline LSSAE 1.1\%.

\begin{figure}[t!]
    \centering
    \includegraphics[width=7.5cm]{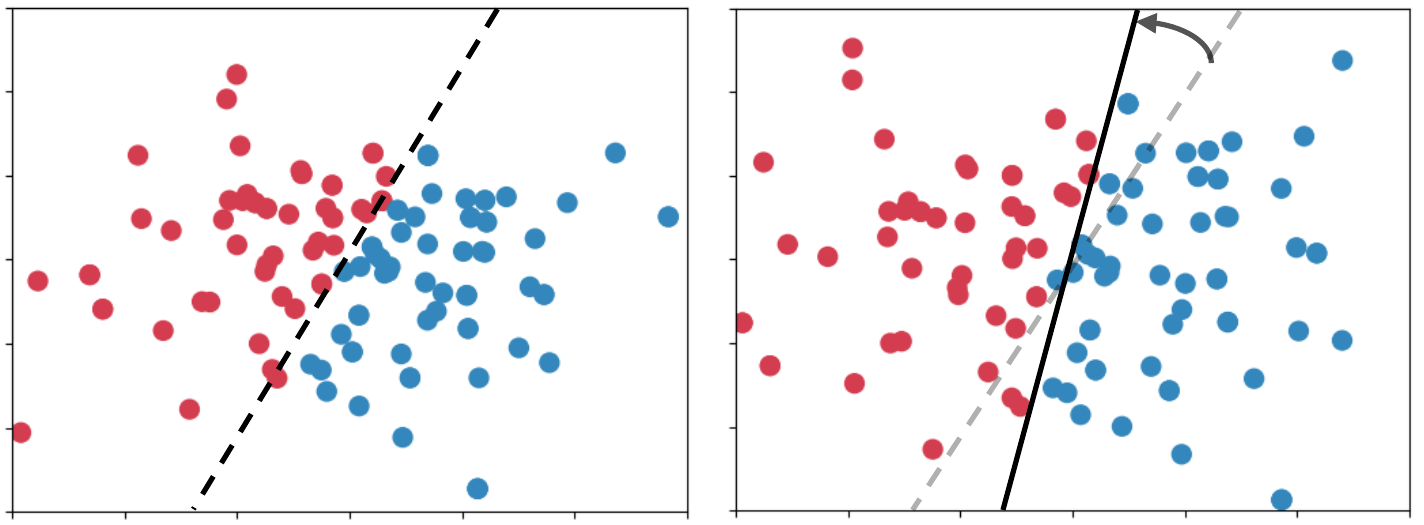}
    \caption{Visualizations of the directional augmentations in rotated Gaussian Datasets. The dashed line represents the decision boundary for data in $T$-th domain. The solid line represents the decision boundary for data in $T+1$-th domain. (Left) Instance embeddings $Z^T$ (Right) Directional transform augmentations $\tilde{Z}^{T+1}$}
    \label{Fig:Rotated gaussian augments}
\end{figure}

\begin{table}[h!]
\setlength{\tabcolsep}{3.5pt}
\begin{center}
\begin{small}
\begin{sc}
\begin{tabular}{cccc}
\toprule
\textbf{interval} &	\textbf{10$^{\circ}$}	&	\textbf{20$^{\circ}$} &	\textbf{30$^{\circ}$}\\ 
\midrule
ERM &  90.2 $\pm$ 0.3 & 75.8 $\pm$ 0.3  & 62.0 $\pm$ 0.2  \\  
MLDG &  92.2 $\pm$ 0.1 & 80.9 $\pm$ 0.3 & 70.6 $\pm$ 0.2 \\  

CIDA & 92.0 $\pm$ 1.2&  85.2 $\pm$ 1.4& 72.1 $\pm$ 1.2\\
EAML & 92.2 $\pm$ 0.5 & 84.7 $\pm$ 0.4 & 71.5 $\pm$ 0.4\\
LSSAE & 92.5 $\pm$ 0.4& 85.5 $\pm$ 0.3& 72.4 $\pm$ 0.4\\
GI & 93.3 $\pm$ 0.2  & 85.3 $\pm$ 0.1 & 71.8 $\pm$ 0.2\\ 
Our Method & \textbf{95.1 $\pm$ 0.2}&  \textbf{86.1 $\pm$ 0.3}& \textbf{73.5 $\pm$ 0.2}\\  

\bottomrule
\end{tabular}
\end{sc}
\end{small}
\end{center}
\caption{Experiment on Rotating MNIST with different intervals with total 9 domains}
\label{Table:Interval}
\end{table}
\subsection{Visualizations of Directional Transform Augmentations}

In Fig.~\ref{Fig:Rotated gaussian augments}, we visualize the augmentations and the source domain data in the rotated Gaussian dataset. From Fig.~\ref{Fig:Rotated gaussian augments}, the decision boundary of $\tilde{Z}^{T+1}$ corresponds to the $T+1$-th domain's. It verifies that DDA successfully generates augmentations which have the same decision boundary as the instances in the next target domain. 

\section{Conclusion}
In this paper, we address the challenging problem of Evolving Domain Generalization. We first show that a specially designed domain transformer learns to capture the domain shifts. Then, we introduce the meta-learning framework to solve the formulated bi-level optimization problem. We conduct extensive experiments on multiple datasets to demonstrate its superior effectiveness. We will further investigate when and how our method solves more complex non-stationary problems in future.

\section*{Ethics Statement}
This paper proposes an algorithm that leverages evolving patterns to make predictions on the unseen target domain. The dataset we use is only intended to demonstrate the algorithm's superior performance on classification tasks.

\section*{Acknowledgements}
We appreciate constructive feedback from anonymous reviewers and meta-reviewers. This work is supported by the Natural Sciences and Engineering Research Council of Canada (NSERC), Discovery Grants program.

\bibliography{egbib}

\appendix

\clearpage

\section{Proof of Equation~\ref{eqn:rearange}}

\begin{proof}
Since 
\label{sup:proof}
\begin{align*}
\|Y^t - X^t \psi  h_{t+1}^*\|_2^2 & =  {h_{t+1}^*}^\mathsf{T}\psi^\mathsf{T} {X^t}^\mathsf{T} X^t \psi  h_{t+1}^* \\
& -2{Y^t}^\mathsf{T}  X^t \psi  h_{t+1}^* + {Y^t}^\mathsf{T} Y^t 
\end{align*}
Setting the above equation derivative with respect to $\psi$ as $0$,  we obtain
\begin{align*}
\frac{\partial \|Y^t - X^t \psi  h_{t+1}^*\|_2^2}{\partial \psi}  =0 ,
\end{align*}
which gives
\begin{align*}
&2{X^t}^\mathsf{T} X^t \psi h_{t+1}^*  {h_{t+1}^*}^\mathsf{T}- 2{X^t}^\mathsf{T} Y^t {h_{t+1}^*}^\mathsf{T}=0  \\
\Rightarrow &{X^t}^\mathsf{T} X^t \psi h_{t+1}^*  {h_{t+1}^*}^\mathsf{T}  = {X^t}^\mathsf{T} Y^t {h_{t+1}^*}^\mathsf{T}.
\end{align*}
Vectorizing both sizes of the equation gives
\begin{align*}
&\biggl(   \bigl( h_{t+1}^*  (h_{t+1}^*)^\mathsf{T}\bigl) \otimes \bigl({X^t}^\mathsf{T} X^t\bigl) \biggl)\text{vec}(\psi) = \text{vec}({X^t}^\mathsf{T} Y^t {h_{t+1}^*}^\mathsf{T})
\end{align*}
\end{proof}

\section{Experiment Datasets}
\label{sec:dataset}
In this subsection, we illustrate the datasets in the experiments. 

\noindent\textbf{Rotating MNIST dataset}~\cite{deng2012mnist} Rotating MNIST is a semi-synthetic dataset where we rotate each MNIST image by a certain angle for a certain domain. In standard setting, we sample 800 digit pictures from MNIST dataset for each domain and rotate them by $0^{\circ}$, $15^{\circ}$, $30^{\circ}$, $45^{\circ}$, $60^{\circ}$ or $75^{\circ}$. The last domain will remain unseen during training and be treated as the test target domain. It contains 70, 000 samples of the dimension (1, 28, 28) with 10 classes. We use MNIST ConvNet architecture as the backbone designed by~\cite{domainbed}.

\noindent\textbf{Rotated Gaussian}~\cite{wang2020continuously} is a synthetic dataset that consists of 30 domains. The instances of each domain are generated by the same Gaussian distribution, but the decision boundary rotates from $0^{\circ}$ to $338^{\circ}$ with an interval of $12^{\circ}$. For every domain, we randomly sample only 125 instances. Fig.~\ref{fig:Rotated gaussian} visualizes this synthetic dataset.

\noindent\textbf{Sine}~\cite{wang2020continuously} includes 11 domains and each domain occupy $\frac{1}{6}$ the period of the sinusoid. We consider the first 10 domains as the source domains and 11 domains as the target domain. For the feature extractor $\phi$, we use the direct identity function because the input size is only 2. The classification model is set to a single linear layer. This setup makes the task very challenging, but our method still could handle it. Fig. ~\ref{fig:Sine} visualize the dataset.

\noindent\textbf{Portraits}~\cite{Ginosar2015ACO} A real dataset consists of photos of high school students across years. The task of the dataset is the binary classification task to classify students' gender (male and female). We divide the dataset into 11 domains along with the years. Each domain has 689 images.

\noindent\textbf{Forest Cover}~\cite{Kumar2020UnderstandingSF} Forest Cover dataset aims to predict cover type (the predominant kind of tree cover) from 54 strictly cartographic variables. To generate non-stationary environments, we sort the samples by the ascending order of the height of the water body, as proposed in ~\cite{Kumar2020UnderstandingSF}. We equally divide the dataset into 10 domains by the altitude (the height of the water body). 

\noindent\textbf{Ocular Disease} (from the Kaggle Competition~\cite{kaggle}) Ocular Disease Intelligent Recognition (ODIR) is a structured ophthalmic database of 5,000 patients with age, color fundus photographs from left and right eyes and doctors' diagnostic keywords from doctors. We set three classes: Normal, Diabetes and other diseases. To generate non-stationary environments, we sort the photographs in ascending order of the age of the patients. We divide the dataset into 10 domains.

\noindent\textbf{Caltran}~\cite{hoffman2014continuous} Caltran is a real-world surveillance dataset consisting of images collected by a traffic camera deployed at an intersection. The task is
to predict the type of scene based on continuously evolving
data. We divide it into 46 domains based on different time periods.

\section{Experiment Setting}
Neural network architectures used for different datasets in Table~\ref{Table:NN arch}. Wide ResNet, ResNet18,and MNIST ConvNet are from domainbed codes~\cite{domainbed}.

\begin{table*}[h!]

\setlength{\tabcolsep}{3.5pt}
\begin{center}
\begin{small}
\begin{sc}
\begin{tabular}{cccc}
\toprule
Dataset & Feature Extractor & Classifier & Domain Transformer \\
\midrule
Sine & Identity Function & A Linear Layer & $[2, 16, 16, 2]$-MLP \\
Rotated Gaussian & Identity Function & A Linear Layer & $[2, 4, 4, 2]$-MLP \\

Portrait & Wide ResNet &  $[128, 64, 32, 2]$-MLP & $[128, 128, 128, 128]$-MLP \\
Rotating MNIST & MNIST ConvNet &  $[128, 64, 32, 10]$-MLP & $[128, 128, 128, 128]$-MLP \\ 
Forest Cover & $[54, 256]$-MLP & $[256, 128, 64, 2]$-MLP & $[256, 256]$-MLP \\ 
Ocular Disease & ResNet-18 & $[512, 256, 128, 3]$-MLP & $[512, 512, 512]$-MLP \\
Caltran & ResNet-18 & $[512, 256, 128, 2]$-MLP & $[512, 512]$-MLP \\

\bottomrule
\end{tabular}
\end{sc}
\end{small}
\end{center}
\caption{Neural network architectures for different datasets. (MLP is short for Multiple-layer Perceptrons , $D$ is the input dimension, and $K$ is the number of classes)}
\label{Table:NN arch}
\end{table*}

We set the default temperature $\tau_{\text{temp}}$ to $2$, the inner-loop update steps to $2$ for source domains, and $5$ for target domains, and list the values of the rest hyper-parameters for different datasets in Table~\ref{Table:hyper}. 

\begin{table}[h!]

\setlength{\tabcolsep}{3.5pt}
\begin{center}
\begin{small}
\begin{sc}
\begin{tabular}{ccc}
\toprule
Dataset & Parameters & Value \\
\midrule
\multirow{3}{*}{Sine} & $\alpha$ & 0.5 \\
& $\beta$ & 0.01\\
& $\lambda$ & 0.8\\
\hline
\multirow{3}{*}{Rotated Gaussian} & $\alpha$ & 2 \\
& $\beta$ & 0.001\\
& $\lambda$ & 0.5\\

\hline
\multirow{3}{*}{Portrait}  & $\alpha$ & 0.05 \\
& $\beta$ & 0.001\\
& $\lambda$ & 0.8\\
\hline
 \multirow{3}{*}{Rotating MNIST}  & $\alpha$ & 0.01 \\
& $\beta$ & 0.005\\
& $\lambda$ & 0.5\\
\hline
\multirow{3}{*}{Forest Cover }  & $\alpha$ & 0.1 \\
& $\beta$ & 0.01\\
& $\lambda$ & 0.5\\
\hline
\multirow{3}{*}{Ocular Disease}  & $\alpha$ & 0.05 \\
& $\beta$ & 0.5 \\
& $\lambda$ & 5$e$-5\\
\hline
\multirow{3}{*}{Caltran}  & $\alpha$ & 0.05 \\
& $\beta$ & 0.5 \\
& $\lambda$ & 5$e$-5\\

\bottomrule
\end{tabular}
\end{sc}
\end{small}
\end{center}
\caption{Hyper-parameters and selected values}
\label{Table:hyper}
\end{table}

\section{Evaluation on Various Experimental Setting on Rotating MNIST}
Except for the standard experiment in Table \ref{Table:real-world}, we further investigate how the experiment setting will influence the performance of our model and baselines. 
In Table~\ref{Table:Interval-supp}, the intervals of rotation degrees between domains are set to 10$^{\circ}$, 15$^{\circ}$, 20$^{\circ}$, 30$^{\circ}$ and the total number of domains is fixed to 9. Our proposed method outperforms all the baselines. As the domain interval increases, all methods' performance degrades. This is because, with a larger domain discrepancy caused by the bigger domain interval, the model gets harder to capture the robust representations for classification tasks. Specifically, when the domain interval is 20$^{\circ}$, our method outperforms the best baseline MLDG~cite 5.2\%.

In Table~\ref{Table:Sample}, we try various numbers of samples per domain. We find with fewer samples per domain, our method improves the baselines with a larger margin. When we set every domain to have 500 samples, our method exceeds the second-best method MLDG by more than 5.4\%. This success of our method indicates that training the model to capture the domain shift patterns can significantly promote its performance on the tasks of Evolving Domain Generalization. It also indicates that the baselines cannot learn invariant representations across domains with a small number of samples in each domain.



\begin{table}[h!]
\setlength{\tabcolsep}{3.5pt}
\begin{center}
\begin{small}
\begin{sc}
\begin{tabular}{ccccc}
\toprule
\textbf{interval} &	\textbf{10$^{\circ}$}&	\textbf{15$^{\circ}$}	&	\textbf{20$^{\circ}$} &	\textbf{30$^{\circ}$}\\ %
\midrule
ERM &  90.2 & 85.5 & 75.8   & 62.0  \\
 GroupDRO & 91.1 & 83.5 & 79.8 & 63.9 \\
 IRM & 75.0 & 67.1 & 55.1 & 48.6 \\
MMD & 88.5 & 82.8 & 75.6 & 45.9  \\ 
 CORAL &91.9 & 84.1 & 77.6 & 63.2 \\
 MTL & 92.8 & 84.1 & 77.5 & 63.1\\
MLDG &  92.2  & 85.9 & 80.9  & 70.6  \\
SagNet &  91.9 & 86.8 & 79.2 & 62.6 \\

SelfReg& 93.0 & 87.5 & 77.9 & 67.5\\
Our Method & \textbf{95.1 }& \textbf{89.1} & \textbf{86.1 }& \textbf{73.5 }\\  

\bottomrule
\end{tabular}
\end{sc}
\end{small}
\end{center}
\caption{Experiment on Rotating MNIST with different intervals with total 9 domains}
\label{Table:Interval-supp}
\end{table}

\begin{table}[h!]
\setlength{\tabcolsep}{3.5pt}
\begin{center}
\begin{small}
\begin{sc}

\begin{tabular}{ccccc}
\toprule
\textbf{Num. Sample} &	200 &	500	& 800 & 2000\\
\midrule
ERM & 64.0 & 70.2 & 75.8 & 83.2 \\
GroupDRO  &  67.5 & 73.0 & 79.8 & 83.9  \\
IRM  &  46.5 & 47.6 & 55.1 & 60.0 \\
MMD  &  71.0 & 73.0 & 75.6 & 82.0 \\ 
CORAL &  66.5 & 70.4 & 77.6 & 82.3\\
MTL & 64.5 & 67.8 & 77.5 & 83.8 \\
MLDG &  71.5 & 73.2 & 80.9 & 85.5 \\
SagNet &  69.5 & 72.8 & 79.2 & 81.5 \\
SelfReg & 71.5 & 72.8 & 77.9 & 84.5 \\
Our Method & \textbf{78.0} & \textbf{78.6} & \textbf{86.1} & \textbf{87.5} \\

\bottomrule
\end{tabular}
\end{sc}
\end{small}
\end{center}
\caption{Experiment on Rotating MNIST with a different number of samples per domain with fixed domain interval 20$^{\circ}$}
\label{Table:Sample}
\end{table}

\section{Modifications to DDA for the setting of multiple target domains}
\label{sup:multi dda}
In the training phase, we take $\tilde{z}_i^{t+2}=\psi(\tilde{z}_i^{t+1})$ ($3 \le t+2 \le T$) as an augmentation to the $(t+2)$-th domain, so $\psi$ gains the ability to generate the augmentation of its next domain. In the test (inference) phase, we use $\tilde{z}_i^{T+2}=\psi (\tilde{z}_i^{T+1})$ as an augmentation of $(t+2)$-th domain and $\tilde{z}_i^{T+3}=\psi(\tilde{z}_i^{T+2})$ as an augmentation of $(T+3)$-th domain. In this way, we could get $\theta_{h_{T+2}}$ and $\theta_{h_{T+3}}$ through fast adaptations on $\{\tilde{z}_i^{T+2}\}_{i=1}^{n_T}, \{\tilde{z}_i^{T+3}\}_{i=1}^{n_T}$.




\section{Ratio in the Distillation Loss}
We try different ratios in the distillation loss in Eqn.~\ref{Eqn:Inner} to evaluate the trade-off between the classification cross-entropy and the distillation loss. We can find that DDA works better with a bigger $\lambda$ and a larger weighting on the hard classification loss. The soft targets can retain the semantic information of the original samples. Therefore, a softened version classification could contribute to the meta-parameter $\theta_h$ adapting to a more robust classification model.

\begin{table}[t!]
\setlength{\tabcolsep}{3.5pt}

\begin{center}
\begin{small}
\begin{sc}
\begin{tabular}{ccccccc}
\toprule
\textbf{$\lambda$} & 0 & 0.2 & 0.4 & 0.6 & 0.8 & 1 \\
\midrule
Our Method &90.7& 91.8 &91.9 & 94.9&  93.3 & 92.0\\

\bottomrule
\end{tabular}
\end{sc}
\end{small}
\end{center}
\caption{Experiment on Ratio in the Distillation Loss on Portrait.}
\label{Table:RatioDistil}
\end{table}


\section{Baselines with Domain Index}
We implicitly use the domain index in the training process. For a fair comparison, we also add the domain index as inputs in the other baseline methods by directly appending the domain index to the embedded features. However, the performance of most baselines degrades and is still worse than our method. We show these experiment results in table~\ref{Table:dm index}. This is counter-intuitive because usually, additional information will improve the task performance. The domain information may need to be used carefully. Otherwise, the domain information will not be helpful to the performance. 

\begin{table}[h!]
\setlength{\tabcolsep}{3.5pt}
\begin{center}
\begin{small}
\begin{sc}
\resizebox{1\columnwidth}{!}{
\begin{tabular}{ccccc}
\toprule
\multirow{2}{*}{\textbf{dataset}}& \multicolumn{2}{c}{\textbf{Rotating MNIST}}&\multicolumn{2}{c}{\textbf{Sine}}\\
 	&  w/ index &	w/o index &  w/ index &	w/o index\\
\midrule
ERM &  74.0 & 75.8 &  57.1 & 56.3	 \\
GroupDRO & 77.8 & 79.8 & 62.4 & 62.6 \\
IRM & 68.6 & 55.1 & 87.7 & 51.1\\
MMD & 75.4 & 75.6 & 54.2 & 54.7\\ 
CORAL &76.9 & 77.6 & 54.9 & 54.7 \\
MTL & 76.8 & 77.5 & 53.6 & 54.2\\ 
MLDG &  80.3 & 80.9 &  54.4 & 54.7  \\
SagNet &  77.9 & 79.2 &  49.9 & 51.1 \\
SelfReg& 77.1 & 77.9 & 56.3 & 55.8 \\
Our Method & \textbf{86.1} & - & \textbf{93.8} &  - \\

\bottomrule
\end{tabular}}
\end{sc}
\end{small}
\end{center}
\caption{Experiment on Rotating MNIST (with interval \textbf{20$^{\circ}$} with total 9 domains) and Sine dataset}
\label{Table:dm index}
\end{table}


\end{document}